\def\year{2021}\relax
\documentclass[letterpaper]{article} 
\usepackage{aaai21}  
\usepackage{times}  
\usepackage{helvet} 
\usepackage{courier}  
\usepackage[hyphens]{url}  
\usepackage{graphicx} 
\usepackage{natbib}  
\usepackage{caption} 
\usepackage{rotating,tabularx}
\usepackage{multirow}
\usepackage{float}
\usepackage{booktabs} 
\usepackage{url}
\usepackage{graphicx}  
\usepackage{amsfonts} 
\usepackage{amsmath}
\usepackage{bm}
\usepackage{float}
\usepackage{multirow}
\usepackage{subfigure}
\usepackage{algorithm}  
\usepackage{CJK}
\usepackage{algpseudocode}  
\usepackage{makecell}
\usepackage{xspace}
\usepackage{amssymb}
\usepackage{array}
\usepackage{footmisc}
\usepackage{array}

\usepackage{booktabs}
\usepackage{listings}
\usepackage[np, autolanguage]{numprint}
\usepackage{paralist}
\usepackage{enumitem}
\usepackage{colortbl}
\usepackage{soul}
\usepackage{epstopdf}
\usepackage{pifont}
\usepackage{xcolor}
\usepackage{mathrsfs}
\usepackage{todonotes}
\usepackage{subfig}

\makeatletter
\newcommand{\thickhline}{%
	\noalign {\ifnum 0=`}\fi \hrule height 1pt
	\futurelet \reserved@a \@xhline
}
\makeatother

\title{Reasoning in Dialog:\\ Improving Response Generation by Context Reading Comprehension}
\author{
    Xiuying Chen\textsuperscript{\rm 1,2}, Zhi Cui\textsuperscript{\rm 3}, Jiayi Zhang\textsuperscript{\rm 3}, Chen Wei\textsuperscript{\rm 3},\\ Jianwei Cui\textsuperscript{\rm 3}, Bin Wang\textsuperscript{\rm 3}, Dongyan Zhao\textsuperscript{\rm 1,2}, and Rui Yan\textsuperscript{\rm 4,5}\thanks{Corresponding author (ruiyan@ruc.edu.cn).}
    \\
}
\affiliations{
    \textsuperscript{\rm 1}Center for Data Science, AAIS, Peking University,Beijing,China\\
    \textsuperscript{\rm 2}Wangxuan Institute of Computer Technology, Peking University,Beijing,China\\
     \textsuperscript{\rm 3}Xiaomi AI Lab\\
        \textsuperscript{\rm 4}Gaoling School of Artificial Intelligence, Renmin University of China\\
         \textsuperscript{\rm 5}Beijing Academy of Artificial Intelligence \\
      xy-chen@pku.edu.cn

}

\begin{document}

\maketitle

\begin{abstract}
In multi-turn dialog, utterances do not always take the full form of sentences \cite{Carbonell1983DiscoursePA}, which naturally makes understanding the dialog context more difficult.
However, it is essential to fully grasp the dialog context to generate a reasonable response.
Hence, in this paper, we propose to improve the response generation performance by examining the model's ability to answer a reading comprehension question, where the question is focused on the omitted information in the dialog.
Enlightened by the multi-task learning scheme, we propose a joint framework that unifies these two tasks, sharing the same encoder to extract the common and task-invariant features with different decoders to learn task-specific features.
To better fusing information from the question and the dialog history in the encoding part, we propose to augment the Transformer architecture with a memory updater, which is designed to selectively store and update the history dialog information so as to support downstream tasks.
For the experiment, we employ human annotators to write and examine a large-scale dialog reading comprehension dataset.
Extensive experiments are conducted on this dataset, and the results show that the proposed model brings substantial improvements over several strong baselines on both tasks. 
In this way, we demonstrate that reasoning can indeed help better response generation and vice versa.
We release our large-scale dataset for further research\footnote{\url{https://github.com/yingtaomj/Reasoning-in-Dialog}}.
\end{abstract}

\section{Introduction}
In recent years, text generation has made impressive progress~\cite{chen2019learning,li2020vmsmo,yu2020draft,liu2020character,zhang2021writing}, and open-domain dialogue generation has become a research hotspot in Natural Language Processing due to its broad application prospect, including chatbots, virtual personal assistants \cite{qiu2019training,Debnath2018IdentifyingC,li2019insufficient}, etc. 
However, studies \cite{Carbonell1983DiscoursePA} show that users of dialogue systems tend to use succinct language which often omits entities or concepts made in previous utterances.
To make appropriate responses, dialogue systems must be equipped with the ability to understand these incomplete utterances.
This naturally leads to the reading comprehension task, where correctly answering questions about the context requires understanding of natural language of the dialog context~\cite{Rajpurkar2016SQuAD10}.

\begin{table*}
	\vspace{-3mm}
	\small
	\renewcommand\arraystretch{0.8}
	\centering
	\setlength{\tabcolsep}{2mm}{
	\begin{tabular}{c|ccc} 
		\hline
		& Example 1 & Example 2  & Example 3      
		\\ 
		\hline
		$A_1$    &
		\begin{tabular}[c]{@{}c@{}}\begin{CJK}{UTF8}{gbsn}求帮忙\textcolor{red}{取}名字姓程，俩男娃\end{CJK}
			\\Please help me decide how to \textcolor{red}{name} \\my two kids whose last name is Cheng\\
			
		\end{tabular}        
		&
		\begin{tabular}[c]{@{}c@{}}\begin{CJK}{UTF8}{gbsn}我最喜欢的歌手是MJ
			\end{CJK}
			\\My favorite singer is MJ\\
			
		\end{tabular}  
		&
		\begin{tabular}[c]{@{}c@{}}\begin{CJK}{UTF8}{gbsn}那么我们即使不死,也在天堂
			\end{CJK}
			\\Then we are in heaven even\\ if we don't die\\
		\end{tabular} 
		\\
		$B_1$     & 
		\begin{tabular}[c]{@{}c@{}}\begin{CJK}{UTF8}{gbsn}程饭和程菜
			\end{CJK}
			\\Cheng fan and Cheng cai\\				
			
		\end{tabular}  &
		\begin{tabular}[c]{@{}c@{}}\begin{CJK}{UTF8}{gbsn}你最喜欢他的什么歌呢？
			\end{CJK}
			\\What's your favorite song?\\				
			
		\end{tabular}
		&
		\begin{tabular}[c]{@{}c@{}}\begin{CJK}{UTF8}{gbsn}这话哪抄的
			\end{CJK}
			\\Where did you copy that\\				
			
		\end{tabular}
		\\ 
		$A_2$   & 
		\begin{tabular}[c]{@{}c@{}}\begin{CJK}{UTF8}{gbsn}哈哈哈哈哈\end{CJK}
			\\ LOL \\  
			
		\end{tabular} & \begin{tabular}[c]{@{}c@{}}\begin{CJK}{UTF8}{gbsn}Thriller\end{CJK}
			\\ Thriller\\  
			
		\end{tabular}
		&
		\begin{tabular}[c]{@{}c@{}}\begin{CJK}{UTF8}{gbsn}三毛
			\end{CJK}
			\\ Sanmao\\  
			
		\end{tabular} 
		\\
		$B_2$    & 
		\begin{tabular}[c]{@{}c@{}}\begin{CJK}{UTF8}{gbsn}请务必接受我的建议\end{CJK} 
			\\Please accept my advice\\  
			
		\end{tabular} 
		& 	\begin{tabular}[c]{@{}c@{}}\begin{CJK}{UTF8}{gbsn}我没听过呢，有\textcolor{blue}{这首歌的mv}吗
			\end{CJK} 
			\\I haven't heard of it. \\Is there \textcolor{blue}{an MV of this song}?\\  
			
		\end{tabular}  & 
		\begin{tabular}[c]{@{}c@{}}\begin{CJK}{UTF8}{gbsn}想起以前豆瓣有个帅哥叫东门\end{CJK} 
			\\Remember that there was a\\ handsome man named Dongmen \\in Douban\\  
			
		\end{tabular}     
		\\
		$A_3$ & 
		\begin{tabular}[c]{@{}c@{}}\begin{CJK}{UTF8}{gbsn}咱俩一起生我就接受\end{CJK}  \\
			\begin{CJK}{UTF8}{gbsn}（取名程饭和程菜）\end{CJK}
			\\I'll accept that (name as Cheng \\fan and Cheng cai) if\\ they are our children
		\end{tabular} 
		& \begin{tabular}[c]{@{}c@{}}\begin{CJK}{UTF8}{gbsn}有(这首歌的MV)，我发给你看
			\end{CJK}  
			\\Yes (I have the MV), I'll send it to you
		\end{tabular}   & 
		\begin{tabular}[c]{@{}c@{}}\begin{CJK}{UTF8}{gbsn}那我（豆瓣）叫个南亭算了
			\end{CJK}  
			\\Then my name (in Douban) \\will be Nanting.
		\end{tabular}     
		\\ 
		\hline
		Question    & \begin{tabular}[c]{@{}c@{}}\begin{CJK}{UTF8}{gbsn}如果一起生娃那孩子\textcolor{red}{叫}什么
			\end{CJK}  \\If the children are ours,\\ how to \textcolor{red}{call} them? \end{tabular}       & 
		\begin{tabular}[c]{@{}c@{}}\begin{CJK}{UTF8}{gbsn}准备发什么？
			\end{CJK}  \\What is going to be sent? \end{tabular}   & 
		\begin{tabular}[c]{@{}c@{}}\begin{CJK}{UTF8}{gbsn}南亭是什么的\textcolor{orange}{ID}？
			\end{CJK}  \\Nanting is \textcolor{orange}{ID} of what \end{tabular}      
		\\
		Answer    & \begin{tabular}[c]{@{}c@{}}\begin{CJK}{UTF8}{gbsn}程饭和程菜
			\end{CJK}  \\Cheng fan and Cheng cai \end{tabular}  & 
		\begin{tabular}[c]{@{}c@{}}\begin{CJK}{UTF8}{gbsn}发thriller的的mv
			\end{CJK}  \\Thriller MV \end{tabular}  & 
		\begin{tabular}[c]{@{}c@{}}\begin{CJK}{UTF8}{gbsn}豆瓣
			\end{CJK}  \\Douban \end{tabular}      \\
		\hline
		$B_3$    & \begin{tabular}[c]{@{}c@{}}\begin{CJK}{UTF8}{gbsn}我觉得这名字很好听啊
			\end{CJK}  \\I think it's a nice name \end{tabular}   & \begin{tabular}[c]{@{}c@{}}\begin{CJK}{UTF8}{gbsn}好啊，我一直想看他的MV呢
			\end{CJK}  \\Good, I've always\\ wanted to see his MV \end{tabular}  & 
		\begin{tabular}[c]{@{}c@{}}\begin{CJK}{UTF8}{gbsn}豆瓣就差你这个ID了
			\end{CJK}  \\Douban is waiting for your ID \end{tabular}     \\
		\hline
		\begin{tabular}[c]{@{}c@{}}Reasoning  \\Type \end{tabular} 
		& \textcolor{red}{Paraphrasing} (49.0\%)
		& \textcolor{blue}{Lexical match} (28.5\%)  & \textcolor{orange}{Pragmatics} (22.5\%)   \\
		\hline
	\end{tabular}}
\caption{Examples from the dataset. Questions are concentrated on the omitted information of $A_3$ (which is shown in brackets), and reasoning type is the type of ability that is needed to answer the question.}
\label{tab:intro}
\end{table*}

Take Example 2 in Table~\ref{tab:intro} for example, contents in parentheses are information omitted in the utterance.
Humans are capable of comprehending such missing utterances dependent based on previous utterances and commonsense.
For instance, $A_3$ means sending an \textit{MV} to B instead of a \textit{gift}.
However, though of high importance, it is difficult for models to capture the implicit dependency between utterances without specific design, and that is why the reading comprehension task is proposed \cite{Rajpurkar2016SQuAD10,reddy2019coqa}.
In this case, by reasoning and correctly answering the question with keyword ``MV'', the model learns that the dialog is focused on MV, which leads to a proper response that is also concentrated on music.
Such cases that require dependency on the previous context to fully comprehend current utterance takes up about 60\% according to a survey in \citet{Pan2019ImprovingOD}.
This inspires us to come up with a multi-task framework that generates the response and answers reading comprehension question at the same time, which can boost the performance of each task.

Our\textit{ Multi-task Response Generator} (MRG) augments the previously proposed Transformer architecture \cite{vaswani2017attention} with the ability to encode multiple utterances in a question-aware fashion.
The proposed model first uses a cross-attention mechanism \cite{vaswani2017attention} between the question and dialog words to identify representative words in dialog with the help of question. 
Concretely, we propose a memory updater, which updates its memory state using both the current inputs and previous memory state.
The memory state can be interpreted as a container of the highly summarized dialog history information. 
During the cross-attention process, the current dialog representation is enhanced with the memory state from the previous step.
MRG then uses a hierarchical inner attention, first over different words in each utterance, and then over all utterances in dialog history, to successively learn the utterance-level features.
Finally, MRG utilizes the utterance-level and question features to select the answer to the question while generating the response words. 


Since there lacks large-scale dialog reading comprehension datasets, we hire an annotation team to construct a dialog reading comprehension dataset (DRCD).
Concretely, based on the \textit{Restoration-200K} dataset proposed by \citet{Pan2019ImprovingOD}, where the omitted word span is annotated by humans, we ask the annotators to write question where the answer is the missing phrase.
We manually construct 10k cases, based on which we train a question generator and leverage the model to construct questions for the rest of the dataset.
We benchmark several classic dialog generation and reading comprehension baselines on DRCD.
We also conduct experiments to show that the proposed model brings substantial improvements over these baselines on both tasks.
In this way, we demonstrate that reasoning can indeed help better response generation and vice versa.

Our contributions can be summarized as follows: 

$\bullet$ We propose the multi-task learning framework, which jointly answers reading comprehension questions and generates a proper response in multi-turn dialog scenario.

$\bullet$ We augment the Transformer architecture with a memory updater, which helps selectively store and update history dialog information.

$\bullet$ We release a large scale dialog reading comprehension dataset.
Experimental results on this dataset demonstrate the effectiveness of our proposed framework.

\section{Related Work}

\paragraph{Multi-turn Dialog.}
In recent years, text generation has made impressive progress~\cite{li2018generating,chan2019stick,Gao2020Meaningful,xie2020infusing}, and multi-turn dialog  model aims to take a message and utterances in previous turns as input and generates a response~\cite{tao2019one,gao2020learning}.
Several works \cite{zhang2019dialogpt,adiwardana2020towards,chan2020selection} simplify the multi-turn dialog into single-turn problem by simply concatenating multiple sentences into one sentence, and utilized the basic Seq2seq based on RNN or Transformer to model long sequence.
To make better use of multi-turn utterances, \citet{xing2017hierarchical} apply hierarchical attention on word-level and utterance-level information.
There also various dialog datasets \cite{lowe2015ubuntu,zhang2018personalizing,welleck2018dialogue,reddy2019coqa}.
However, these datasets do not contain reading comprehension question-answering pairs.

\paragraph{Machine Reading Comprehension.} 
Machine reading comprehension (MRC) focuses on modeling semantic matching between a question and a reference document, which read the full text to select relevant text spans and then infer answers.
\citet{choi2017coarse} propose hierarchical coarse-to-fine methods in order to mimic the reading mode of human.
\citet{huang2017fusionnet} come up with a fully-aware fusion attention mechanism and apply it on MRC tasks.
Large-scale datasets for MRC have also been proposed in parallel.
CommonsenseQA \cite{talmor2018commonsenseqa} is a dataset for commonsense question answering extracted from CONCEPTNET \cite{speer2016conceptnet}.
DROP \cite{dua2019drop} and COSMOS \cite{huang2019cosmos} focus on factual understanding and commonsense comprehension, respectively.
In this paper, we propose another MRC dataset focused on machine comprehension on dialog corpus.

\paragraph{Multi-task Learning.}
Multi-task learning (MTL) is a learning paradigm in machine learning and it aims to leverage useful information contained in multiple related tasks to help improve the generalization performance of all the tasks\cite{caruana1997multitask}. 
There are a large quantity of natural language processing tasks based on multi-task learning, such as word segmentation, POS tagging, dependency parsing, and text classification \cite{bohnet2012transition,hatori2012incremental,li2013joint,liu2016recurrent}.
\citet{collobert2008unified} describe a single convolutional network that jointly trained several NLP tasks, such as part-of-speech tags, chunks, named entity tags, semantic roles.
\citet{liu2015representation} develop a multi-task deep neural network combining tasks of multiple-domain classification and information retrieval to learn representations across multiple tasks.
In this work, we apply multi-task learning on response generation and reading comprehension on dialog.


\section{Problem Formulation}
\label{sec:formulation}

Before presenting our approach for the dialog reading comprehension multi-task, we first introduce our notations and key concepts. 

We assume that a conversation is conducted between two users. 
Suppose there are already $N^u$ turns in a dialogue, so we have historical utterances as $X = (X_1, X_2, \ldots, X_{N^{u}})$, where each utterance $X_{j}$ is depicted as $X_j=(x^j_1,x^j_2,\ldots,x^j_{N_j})$ and $x^j_i$ denotes a word.
Accordingly, MRG aims to predict the ($N^u$+$1$)-th utterance ,\textit{i.e.}, the response, $Y=(y_1,y_2,\ldots,y_{N^{y}})$, according to the historical utterances $X$:
\begin{equation}
p(Y| X)=\textstyle \prod_{i=1}^{N^{y}} p\left(y_{i} | X,  y_{1}, \ldots, y_{i-1}\right)
\end{equation}
Apart from the reponse generation, we also design a question-answering task for the model.
That is, targeted at the $N^u$-th utterance, where some keywords are missing, there is a question $Q=(q_1,q_2,...,q_{N^q})$ that asks about such missing information, and the answer is a score vector $A=(a_1,a_2,...a_{N^a})$ that extracts the missing keywords from previous utterances. $N^a=\textstyle \sum_{i=1}^{N^u}N_i$.
Each score $a_i \in \{0,1\}$ denotes whether the $i$-th word is selected (1) or not (0).
The objective is to maximize the likelihood of all word labels $A$ given the input:
\begin{equation}
p\left(A| X \right)=\textstyle \prod_{i=1}^{N^a} p\left(a_{i} | X\right)
\end{equation}

\section{The Proposed MRG Model}

\subsection{Overview}

In this section, we propose the \emph{Multi-task Response Generator}, abbreviated as MRG. 
An overview of MRG is shown in Figure~\ref{fig:reasoning-dialog}, which can be split into three main parts:

\begin{figure*}
	\centering
	\includegraphics[width=1\linewidth]{"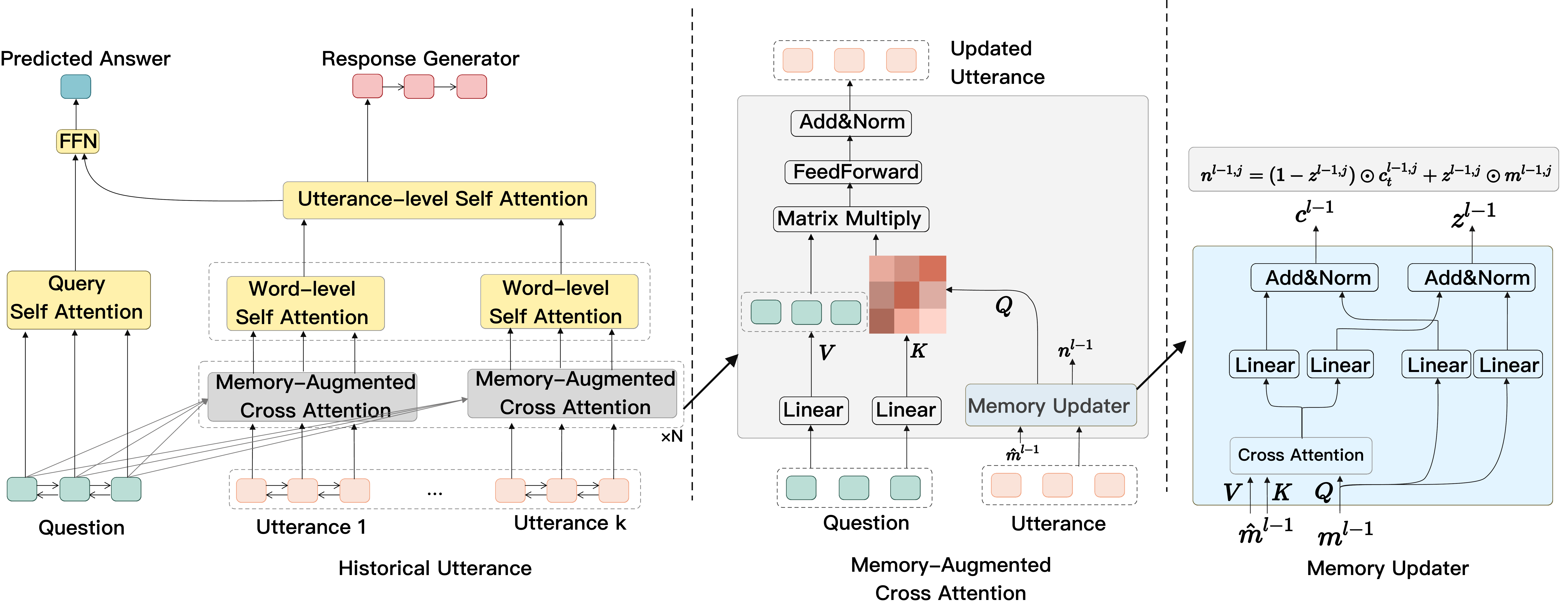"}
	\caption{Overview of MRG. We divide our model into three parts: (1) Cross-hierarchical Encoder (which consists of memory-augment cross attention and two hierarchical self attentions); (2) Answer Selecter; (3) Response Generator.}
	\label{fig:reasoning-dialog}
\end{figure*}

$\bullet$ \textit{Cross-hierarchical encoder} first uses a memory-augmented cross-attention mechanism \cite{vaswani2017attention} between the question and dialog words to identify representative words in dialog with the help of question.
It then uses a hierarchical inner attention, first over different words in an utterance, and then over all utterances in dialog history, to successively learn the utterance-level features.

$\bullet$ \textit{Answer selecter} takes the question representation and utterance-level dialog features as input to predict the answer.

$\bullet$ \textit{Response generator} produces the response by attending to the utterance-level features.

\subsection{Cross-hierarchical Encoder}
To begin with, we use an embedding matrix $e$ to map a one-hot representation of each word in $X$, $Q$, into a high-dimensional vector space. 
We then employ a bi-directional recurrent neural network (Bi-RNN) to model the temporal interactions between words:
\begin{equation}
\begin{aligned}
h_{i}^{x,j}&=\text{Bi-RNN}_{x}\left(e(x_i^j), h_{i-1}^{x,j}\right),\\
h_{i}^{q}&=\text{Bi-RNN}_{y}\left(e(q_i), h_{i-1}^{q}\right),
\end{aligned}
\end{equation}
where $h_{i}^{x,j}$ and $h_{i}^{q}$ denote the hidden state of $i$-th step in Bi-RNN for $X_j$ and $Q$, respectively. 
Following \cite{Zhao2017LearningDD,Chen2018IterativeDR}, we choose long short-term memory (LSTM) as the cell for Bi-RNN.

\paragraph{Memory-augmented Cross Attention.} 
This module grounds the conversation context by the question and fuses the information of the question into the dialog representation.
Concretely, it has a stack of $L$ identical layers. 
In each layer, we iteratively fuse the information from question words to the dialog words by Memory-augmented Cross Attention Module (MCAM). 
For convenience, we denote the output of $l$-th encoder layer as $m^{l,j}_i$ and the input for the first layer $m^{0,j}_i$ is initialized as $h^{x,j}_i$.
Concretely, MCAM is based on the traditional Cross Attention Module (CAM) Transformer architecture \cite{vaswani2017attention}.
We first introduce the original CAM, and then introduce our modification.

The first input for CAM is for query $Q$ and the second input is for keys $K$ and values $V$ for attention, which we denote as $x_i$ and $h^q_*$ respectively:
\begin{equation}
\label{equ:cam}
m^{l,j}_i=\text{CAM}(m^{l-1,j}_i,h^q_{*}).
\end{equation}
Each output element, $m^{l,j}_i$, is computed as weighted sum of a linearly transformed input values:
\begin{equation}
m^{l,j}_i = \textstyle \sum_{k=1}^{N_j} \alpha_{i,k}^{l,j}  \left(h^q_kW^V\right).\label{equ:transformer-sum}
\end{equation}
Each weight coefficient, $\alpha_{i,k}^{l,j} $ , is computed using a softmax function:
\begin{equation}
\alpha_{i,k}^{l,j} =\frac{\exp \left(\beta_{i,k}^{l,j}\right)}{\sum_{k=1}^{N_j} \exp  \left(\beta_{i,k}^{l,j}\right)}.
\end{equation}
And $\beta_{i,k}^{l,j}$ is computed using a compatibility function that compares two input elements:
\begin{equation}
\beta_{i,k}^{l,j}=\frac{\left(m^{l-1,j}_{i} W^{Q}\right)\left(h^q_k W^{K}\right)^{T}}{\sqrt{d}}\label{eq:alpha},
\end{equation}
where $d$ stands for hidden dimension.
$W^Q, W^K, W^V \in \mathbb{R}^{N_j \times N_j}$ are parameter matrices.

While the aforementioned vanilla CAM is a powerful method, it is less suitable for multi-turn dialog due to its inability to fully utilize dialog history information.
Thus, we augment it with an external memory module, which helps to remember and update history dialog information in a multi-slot way as illustrated in Figure~\ref{fig:reasoning-dialog}.
The input for query, \textit{i.e.}, $m^{l-1,j}_i$ is updated to $n^{l-1,j}_i$ through a memory updator, which will then be fed into CAM in Equation~\ref{equ:cam}.
Concretely, the memory updator aggregates the information from both its intermediate hidden states $\hat{m}^{l-1}_i$ ($\hat{m}^{l-1} \in \mathbb{R}^{N_j\times d}$) and the utterance (memory) states $m^{l-1,j}_i$ from the last layer, using a multi-head attention. 
Specifically, the input for query $Q$ is $m^{l-1,j}_i$, and input for key $K$ and value $V$ is $[\hat{m}^{l-1}_i;m^{l-1,j}_i]$. 
The memory augmented hidden states are further encoded using a feed-forward layer and then merged with the intermediate hidden states $\hat{m}^{l-1}$ using a residual connection and layer norm. 
We summarize the procedure below:
\begin{align*}
s^{l-1,j}_i &= \textrm{CAM}(m^{l-1,j}_i, \hat{m}^{l-1}_i), \\
c^{l-1,j}_i &= \textrm{tanh}(W_{a}^{l-1} m^{l-1,j}_i + W_{b}^{l-1} s^{l-1,j}_i), \\
z^{l-1}_i &= \textrm{sigmoid}(W_{c}^{l-1,j}m^{l-1,j}_i + W_{d}^{l-1} s_i^{l-1}), \\
n^{l-1,j}_i &= (1 - z_i^{l-1,j}) \odot c_{t}^{l-1,j} + z^{l-1,j}_i \odot m^{l-1,j}_i,
\end{align*}
where $\odot$ denotes Hadamard product, $W_{a}^{l-1}$, $W_{b}^{l-1}$, $W_{c}^{l-1}$, and $ W_{d}^{l-1}$ are trainable weights, $c_{i}^{l-1}$ is the internal cell state. 
$z_{i}^{l-1}$ is the update gate that controls which information to retain from the previous memory state. 


\paragraph{Hierarchical Self Attention.}
After utilizing question information to emphasize important keywords, $m^{L,j}_i$ (the output of last MCAM layer) is then processed by a hierarchical attentive module to encode long-term dependency among words into the representations.
The first level in our hierarchical attention encodes each utterance independently from other utterances at word-level, resulting in a fixed-dimensional representation of each utterance. 
Concretely, the word-level attentive module simplifies the Multi-head Attention Module (MAM) in Transformer, which is similar to CAM, but takes the same input for query, key and value:
\begin{equation}
h^{w,j}_i=\text{MAM}(m^{L,j}_i,m^{L,j}_*).
\end{equation}
A mean-pooling operation is then used over word vectors in each utterance to obtain a fixed-length utterance-level representation:
\begin{equation}
h^{u',j}=\operatorname{meanpool}\left(\left\{h^{w,j}_1, \cdots, h^{w,j}_{N_j}\right\}\right).
\end{equation}
Similar to word-level attention, an utterance-level MAM is applied on these representations to fuse information between different utterances:
\begin{equation}
h^{u,j}=\text{MAM}(h^{u',j},h^{u',*}).
\end{equation}

From the utterance representation, we can also obtain the overall dialog history representation, which will be used in the response decoder part:
\begin{equation}
h^{d}=\operatorname{meanpool}\left(\left\{h^{u,1}, \cdots, h^{u,N^u}\right\}\right).
\end{equation}

\subsection{Answer Selector} 

After fusing information from question and dialog context, it is time to select words from context as the answer to the question.
Since we have several utterance representations, and either taking the average or summing them together by specific weights is inappropriate and inelegant. 
Hence, we concatenate all utterance and question representations together and apply a multi-layer perceptron to them to generate the word extracting probabilities:

\begin{align*}
h^q&=\operatorname{meanpool}\left(\left\{h^{q}_1, \cdots, h^{q}_{N^q}\right\}\right),\\
\hat{A}&=W_f \tanh \left(W_e\left[h^{u,1} ; \ldots ; h^{u,N^u};h^q\right]+b^e\right)+b^f,
\end{align*}
where $[;]$ denotes concatenation operation.

\subsection{Response Generator} 
To generate a consistent and informative response, we propose an RNN-based decoder that incorporates outputs of utterance representations as illustrated in Figure~\ref{fig:reasoning-dialog}.

We first apply a linear transform layer on the input document vector representation $h^d$ and use the output of this layer as the initial state of decoder LSTM, shown in Equation~\ref{equ:dec-init}.
In order to reduce the burden of compressing document information into the initial state $s_0$, we use the attention mechanism~\cite{Bahdanau2014NeuralMT} to summarize the utterance representations into context vector $f_{t-1}$ dynamically and we will show the detail of these in this section later.
We then concatenate the context vector $f_{t-1}$ with the embedding of previous step output $e(y_{t-1})$ and feed this into decoder LSTM, shown in Equation~\ref{equ:dec-step}:
\begin{align}
s_0 &= W_g h^d + b_g, \label{equ:dec-init}\\
s_t &= \text{LSTM} \left( s_{t-1}, [f_{t-1}; e(y_{t-1})] \right) . \label{equ:dec-step}
\end{align}

Context vector $f_{t-1}$ is the vector that stores the dialog context information at $t$-th step.
Concretely, we use the decoder state $s_{t-1}$ to attend to each utterance states $h^{u,i}$ and results in the attention distribution $\gamma_{t}$, shown in Equation~\ref{equ:attention-sm}.
Then we use the attention distribution $\gamma_{t}$ to weighted sum the document states as the context vector $f_{t-1}$.
\begin{align}
\gamma^{'}_{t-1, i} &= W_n^\intercal \tanh \left( W_s s_{t-1} + W_h h^{u,i} \right), \\ 
\gamma_{t-1, i} &= \exp \left( \gamma^{'}_{t-1, i} \right) / \textstyle \sum^{N_u}_{j=1} \exp \left(\gamma^{'}_{t-1, j} \right), \label{equ:attention-sm}\\
f_{t-1} &= \textstyle \sum_{i=1}^{N_u} \gamma_{t-1, i} h^{u,i} .
\end{align}
Finally, an output projection layer is applied to get the final generating distribution $P_t^{v}$ over vocabulary, as shown in Equation~\ref{equ:out-proj}.
We concatenate utterance context vector and the output of decoder LSTM $s_t$ as the input of the output projection layer:
\begin{equation}
P^v_t = \text{softmax} \left( W_v [s_t; f_t]  + b_v \right), \label{equ:out-proj}
\end{equation}

We use the negative log-likelihood as the loss function:
\begin{align}
\mathcal{L}_g &= - \textstyle \sum^{N^y}_{t=1} \log P_t^{v}(y_t).
\end{align}

\section{Experimental Setup}
\label{sec:experiment}

\subsection{Dataset}
To our best knowledge, no exsiting works consider MRC in response generation task.
Hence, we first propose a dialog reading comprehension dataset (DRCD).
DRCD is based on the \textit{Restoration-200k} dataset proposed by \citet{Pan2019ImprovingOD}, where the utterance with omitted information is manually annotated.
Such omitted information leads to a difficulty in fully understanding the dialog context and requires reasoning ability to  for a model.
Hence, we hire an annotation team to write questions that are focused on the missing information.

Since it is time-consuming to write questions for the whole dataset, and based on the labeled answer it is rather easy to construct the question, we ask the team to write questions for 10k cases, and then automatically generate questions for the rest of the dataset.
Concretely, we utilize PG~\cite{See2017GetTT} to generate questions due to its good performance in many tasks including summarization and dialog completion~\cite{Pan2019ImprovingOD,chen2019learning}.
We then conduct a human evaluation to examine the generation quality.
Concretely, we randomly sample 200 cases and asked three annotators to state how well they agree with the following two statements, on a scale of one to five (strongly disagree, disagree, neutral, agree, or strongly agree):
1) The generated question asks about the omitted phrase.
2) The generated question is written in fluent Chinese.
The result shows that generated questions that score over 3 takes up 76.5\%, showing that most of the generated questions are of good quality.
The kappa statistics indicate the moderate agreement between annotators.

We randomly split the dataset with question-answer pair to 113,116 training, 3,000 validation, and 3,000 test cases.
The average character-level context length and utterance length of the dataset is and 43.4 and 9.05.
Note that in the validation and test datasets the questions are all written by human, ensuring that the testing results are convincing.


\begin{table*}
	\small
	\centering
	{\begin{tabular}{cccccccc}
			\toprule
			\textbf{Model}&\textbf{BLEU1}&\textbf{BLEU2}&\textbf{BLEU3}&\textbf{BLEU4}&\textbf{Average}&\textbf{Extrema}&\textbf{Greedy}\\
			\midrule
			Seq2Seq&0.2260&0.1566&0.0876&0.0671&0.4341&0.6695&0.7759\\
			HRED&0.2273&0.1559&0.0871&0.0667&0.4320&0.6601&0.7885\\
			VAE&0.2316&0.1586&0.0886&0.0680&0.4350&0.6396&0.7808\\
			Transfomer&0.2181&0.1482&0.0825&0.0631&0.4407&0.6500&0.7920\\
			PAC&0.2413&0.1624&0.0902&0.0689&0.4396&0.6447&0.7909\\
			\midrule
			MRG&\textbf{0.2632}&\textbf{0.1735}&\textbf{0.0968}&0.0741&\textbf{0.4513}&\textbf{0.6769}&\textbf{0.8025}
			\\
			\midrule
			MRG w/o MCAM&0.2224&0.1533&0.0857&0.0656&0.4436&0.6630&0.7837\\
			MRG w/o MAM& 0.2404&0.1616&0.0946&0.0665&0.4343&0.6740&0.7798\\
			MRG w/o MemUpd& 0.2498&0.1585&0.0894&\textbf{0.0747}&0.4419&0.6551&0.7884\\
			MRG w/o MT&0.2231&0.1541&0.0862&0.0661&0.4343&0.6734&0.7645\\
			\bottomrule
	\end{tabular}}
\caption{Automatic evaluation results on response generation task. The best results are bold.}
	\label{tab:automated_evaluation}
\end{table*}

\begin{table}
	\small
	\begin{centering}
		\begin{tabular}{lcc}
			\hline 
			\multirow{2}{*}{{Model}} & \multicolumn{2}{c}{{Accuracy(\%)}}\tabularnewline
			& {Mean } & {Best}\tabularnewline
			\hline 
			{MemN2N } & {37.85 } & {38.22}\tabularnewline
			{DMN  } & {40.83 } & {42.21}\tabularnewline
			{QRN } & {40.80} & {43.71}\tabularnewline
			{DMN+ } & {43.97 } & {45.02}\tabularnewline
			\hline 
			{MRG } & \textbf{45.43} & \textbf{47.17}\\
			{MRG w/o MT } & 44.89 & 46.34\tabularnewline
			\hline 
		\end{tabular}
		\par\end{centering}
	\caption{Automatic evaluation results on MRC task. Best accuracy over 10 runs.\label{tab:Results-for-bAbI}}
\end{table}

\subsection{Comparison Methods}

To evaluate the performance of our proposed model, we compare it with the following response generation and MRC baselines:

\noindent \textbf{\textit{Response Generation baselines}}:	

\textbf{Seq2Seq}~\cite{Bahdanau2015NeuralMT}: the vanilla schema of the sequence to sequence model with attention mechanism.

\textbf{HRED}~\cite{Serban2016BuildingED}: extends the hierarchical recurrent encoder-decoder neural network to the dialogue domain.

\textbf{VAE}~\cite{Zhao2017LearningDD}: uses latent variables to learn a distribution over potential conversational intents and generates diverse responses.

\textbf{Transformer}~\cite{vaswani2017attention}:  is based solely on attention mechanisms.

\textbf{PAC}~\cite{Pan2019ImprovingOD}: is a ``pick-and-combine'' model to restore the incomplete utterance from its context, and then use the restored utterance to generate the next response.

\noindent \textbf{\textit{MRC baselines}}:

\textbf{MemN2N}~\cite{Sukhbaatar2015EndToEndMN}: is an extension of RNNsearch to the case with multiple computational hops.

\textbf{DMN}~\cite{Kumar2016AskMA}:  processes input sequences and questions, forms episodic memories, and generates relevant answers.

\textbf{DMN+}~\cite{Xiong2016DynamicMN}: proposes several improvements to memory and input modules of DMN.

\textbf{QRN}~\cite{Seo2017QueryReductionNF}: is a variant of RNN that effectively handles both short-term (local) and long-term (global) sequential dependencies to reason over multiple facts.

\subsection{Implementation Details}

We implement our experiments in TensorFlow~\cite{Abadi2016TensorFlowAS} on an NVIDIA GTX 1080 Ti GPU. 
We truncate input dialog to 100 words, with 20 words in each utterance. 
We chose 100 as our truncation size as we did not find significant improvement when increasing input length from 100 to 200 tokens.
The minimum decoding step is 10, and the maximum step is 20.
The word embedding dimension is set to 128 and the number of hidden units is 256.
We initialize all of the parameters randomly using a Gaussian distribution.
The batch size is set to 16, and we limit the vocabulary size to 50K.
We use Adagrad optimizer~\cite{Duchi2010AdaptiveSM} as our optimizing algorithm.
We also apply gradient clipping~\cite{Pascanu2013OnTD} with a range of $[-2,2]$ during training. 
During the inference stage, the checkpoint with smallest validation loss is chosen and the beam-search size is set to 4 for all methods.
Note that when evaluating the response generation performance, we use the generated questions as input instead of the ground truth human-written questions for the sake of fairness.

\subsection{Evaluation Metrics}
To evaluate the results of the generated responses, we adopt the following metrics widely used in existing research.

\textbf{Overlap-based Metric.} 
Following \citet{Li2021Style,xu2020learning}, we utilize BLEU score \cite{papineni2002bleu}, an algorithm which has been widely used in machine translation and dialogue system, to measure n-grams overlaps between ground-truth and generated response. 
Specifically, we follow the conventional setting in previous work \cite{gu2018dialogwae} to compute BLEU scores using smoothing techniques (smoothing 7).

\begin{table*}
	\small
	\renewcommand\arraystretch{0.8}
	\centering
	\setlength{\tabcolsep}{2mm}{
		\begin{tabular}{c|ccc} 
			\hline
			& Example 1 & Example 2  & Example 3      
			\\ 
			\hline
			$A_1$    &
			\begin{tabular}[c]{@{}c@{}}\begin{CJK}{UTF8}{gbsn}有木有人带妹子吃喝玩乐在杭州	\end{CJK}
				\\Is there anyone to take girls to eat,\\ drink and have fun in Hangzhou\\
				
			\end{tabular}        
			&
			\begin{tabular}[c]{@{}c@{}}\begin{CJK}{UTF8}{gbsn}天蝎座不自恋真的就能死	
				\end{CJK}
				\\Scorpio will die without narcissism\\
				
			\end{tabular}  
			&
			\begin{tabular}[c]{@{}c@{}}\begin{CJK}{UTF8}{gbsn}lz女坐标杭州心情不nice	
				\end{CJK}
				\\The author is a girl, located in \\Hangzhou, has a bad mood\\
			\end{tabular} 
			\\
			$B_1$     & 
			\begin{tabular}[c]{@{}c@{}}\begin{CJK}{UTF8}{gbsn}没真相不敢带
				\end{CJK}
				\\I dare not bring a girl without a photo\\				
				
			\end{tabular}  &
			\begin{tabular}[c]{@{}c@{}}\begin{CJK}{UTF8}{gbsn}挺准最近就是被一个天蝎虐
				\end{CJK}
				\\That's right. I've been\\ abused by a Scorpio recently. \\				
				
			\end{tabular}
			&
			\begin{tabular}[c]{@{}c@{}}\begin{CJK}{UTF8}{gbsn}那怎么办
				\end{CJK}
				\\Then what to do\\				
				
			\end{tabular}
			\\ 
			$A_2$   & 
			\begin{tabular}[c]{@{}c@{}}\begin{CJK}{UTF8}{gbsn}有了真相更不敢带哈	\end{CJK}
				\\ With a photo, you will\\ dare not bring her more \\  
				
			\end{tabular} & \begin{tabular}[c]{@{}c@{}}\begin{CJK}{UTF8}{gbsn}嗯\end{CJK}
				\\ Yes\\  
				
			\end{tabular}
			&
			\begin{tabular}[c]{@{}c@{}}\begin{CJK}{UTF8}{gbsn}睡觉么
				\end{CJK}
				\\ What about sleep\\  
				
			\end{tabular} 
			\\
			$B_2$    & 
			\begin{tabular}[c]{@{}c@{}}\begin{CJK}{UTF8}{gbsn}犀利	\end{CJK} 
				\\Sharp\\  
				
			\end{tabular} 
			& 	\begin{tabular}[c]{@{}c@{}}\begin{CJK}{UTF8}{gbsn}你是摩羯啊
				\end{CJK} 
				\\So you are a Capricorn\\  
				
			\end{tabular}  & 
			\begin{tabular}[c]{@{}c@{}}\begin{CJK}{UTF8}{gbsn}也可以不过不是长久之计	\end{CJK} 
				\\That's good, but not a long-term solution\\  
				
			\end{tabular}     
			\\
			$A_3$ & 
			\begin{tabular}[c]{@{}c@{}}\begin{CJK}{UTF8}{gbsn}一般(犀利)啦\end{CJK}  
				\\Generally (sharp)
			\end{tabular} 
			& \begin{tabular}[c]{@{}c@{}}\begin{CJK}{UTF8}{gbsn}(摩羯被天蝎虐)这就是宿命
				\end{CJK}  
				\\This (Capricorn abused \\by Scorpio) is fate.
			\end{tabular}   & 
			\begin{tabular}[c]{@{}c@{}}\begin{CJK}{UTF8}{gbsn}哈哈那(心情不nice)怎么办
				\end{CJK}  
				\\Ha-ha, then what to do (if in bad mood)
			\end{tabular}     
			\\ 
			\hline
			Question    & \begin{tabular}[c]{@{}c@{}}\begin{CJK}{UTF8}{gbsn}什么很一般
				\end{CJK}  \\What is general \end{tabular}       & 
			\begin{tabular}[c]{@{}c@{}}\begin{CJK}{UTF8}{gbsn}\textcolor{blue}{什么是宿命}
				\end{CJK}  \\\textcolor{blue}{What is fate} \end{tabular}   & 
			\begin{tabular}[c]{@{}c@{}}\begin{CJK}{UTF8}{gbsn}什么出问题了
				\end{CJK}  \\What's wrong \end{tabular}      
			\\
			Answer    &  \begin{tabular}[c]{@{}c@{}}\begin{CJK}{UTF8}{gbsn}犀利程度一般
				\end{CJK}  \\The degree of sharpness \end{tabular} & 
			\begin{tabular}[c]{@{}c@{}}\begin{CJK}{UTF8}{gbsn}摩羯被天蝎虐
				\end{CJK}  \\Capricorn abused by Scorpio \end{tabular}  & 
			\begin{tabular}[c]{@{}c@{}}\begin{CJK}{UTF8}{gbsn}心情
				\end{CJK}  \\Mood \end{tabular}      \\
			\hline
			Reference    & \begin{tabular}[c]{@{}c@{}}\begin{CJK}{UTF8}{gbsn}一般都这么犀利男人咋办	
				\end{CJK}  \\If that is general sharp then \\what to do with man \end{tabular}   & \begin{tabular}[c]{@{}c@{}}\begin{CJK}{UTF8}{gbsn}所以我喜欢上天蝎就是个错误
				\end{CJK}  \\So it was a mistake for me\\ to like Scorpio \end{tabular}  & 
			\begin{tabular}[c]{@{}c@{}}\begin{CJK}{UTF8}{gbsn}找个心灵的驿站
				\end{CJK}  \\Find a post station for the soul \end{tabular}     \\
			\hline
			HRED
			& \begin{tabular}[c]{@{}c@{}}\begin{CJK}{UTF8}{gbsn}是你的错觉
			\end{CJK}\\It's your delusion \end{tabular} &\begin{tabular}[c]{@{}c@{}}\begin{CJK}{UTF8}{gbsn}洗洗睡了
		\end{CJK}\\Wash and sleep \end{tabular} & \begin{tabular}[c]{@{}c@{}}\begin{CJK}{UTF8}{gbsn}生命健康更加可贵
	\end{CJK}\\Life and health are more valuable \end{tabular}\\  
		VAE
		& \begin{tabular}[c]{@{}c@{}}\begin{CJK}{UTF8}{gbsn}好吧,改天可以一起去嗨皮了
			\end{CJK}\\Well, we can play the other day \end{tabular} &\begin{tabular}[c]{@{}c@{}}\begin{CJK}{UTF8}{gbsn}我帮同事发帖子勾搭上了
			\end{CJK}\\I helped my colleagues\\ post and hook up \end{tabular} & \begin{tabular}[c]{@{}c@{}}\begin{CJK}{UTF8}{gbsn}年龄大了就想睡
			\end{CJK}\\When you get older, you want to sleep \end{tabular}\\ 
		PAC
		& \begin{tabular}[c]{@{}c@{}}\begin{CJK}{UTF8}{gbsn}你也要去对不啦
			\end{CJK}\\So you will go there, aren't you? \end{tabular} &\begin{tabular}[c]{@{}c@{}}\begin{CJK}{UTF8}{gbsn}那你还不如玩手机
			\end{CJK}\\You might as well\\ play with your cell phone \end{tabular} & \begin{tabular}[c]{@{}c@{}}\begin{CJK}{UTF8}{gbsn}我闲了就读书
			\end{CJK}\\I read when I'm free \end{tabular}\\ 
		MRG
		& \begin{tabular}[c]{@{}c@{}}\begin{CJK}{UTF8}{gbsn}\textcolor{red}{是过于犀利啦}
			\end{CJK}\\\textcolor{red}{It's too sharp} \end{tabular} &\begin{tabular}[c]{@{}c@{}}\begin{CJK}{UTF8}{gbsn}还是要看人，\textcolor{blue}{怎么就宿命了}哈哈哈
			\end{CJK}\\It still depends on people. \\
		\textcolor{blue}{You should not believe in fate} lol. \end{tabular} & \begin{tabular}[c]{@{}c@{}}\begin{CJK}{UTF8}{gbsn}出门旅游，逛逛豆瓣
			\end{CJK}\\Travel outside, visit Douban \end{tabular}\\ 
			\hline
	\end{tabular}}
	\caption{Responses generated by baselines and our model along with the QA pairs.}
	\label{tab:case}
\end{table*}

\textbf{Embedding Metrics.} To capture the semantic matching degrees between generated responses and ground-truth, we perform evaluations on embedding space. In consistent with previous study \cite{gu2018dialogwae}, we compute the similarity between the bag-of-words (BOW) embeddings representations of generated results and reference.
In particular, we calculate three metrics:1) \textit{Greedy} (BOW-Greedy), i.e., greedily matching words in two utterances based on the cosine similarities; 
2) \textit{Average} (BOW-Average), cosine similarity between the averaged word embeddings in the two utterances \cite{mitchell2008vector};
3) \textit{Extrema} (BOW-Extrema), cosine similarity between the largest extreme values among the word embeddings in the two utterances \cite{forgues2014bootstrapping}.

\textbf{Human Metrics.}
We also employ human evaluation to assess the responses generated by our model and the baselines. 
Three well-educated annotators are hired to evaluate the quality of generated responses, where the evaluation is conducted in a double-blind fashion.
Totally, 200 randomly sampled responses generated by each model are rated by each annotator with two different aspects, \textit{i.e.,} readability (Is the response grammatically formed and smooth?), informativeness (Does the response contains informative words?). 
Criteria are scored from 1 to 3, \textit{i.e.,} bad, normal, and good.

\section{Experimental Results}

\subsection{Overall Performance}
\textbf{Automatic evaluation.}
We first examine whether our MRG outperforms generative baselines as listed in Table~\ref{tab:automated_evaluation}.
Our model outperforms baselines on all automatic metrics. 
This demonstrates that our model generates more appropriate responses by reading comprehension, and understands the dialog context better by predicting response. 
Especially, our model improves approximately 16.46\% over seq2seq on BLEU1, and outperforms PAC by 9.07\%. 
We also list the results of ablation study in Table~\ref{tab:automated_evaluation},
aiming to assess the contribution of individual components. 
Our experiments confirmed that interacting between dialog and question by Memory-augmented Cross Attention Module is beneficial (see row w/o MCAM), as well as self-attention module (see row w/o MAM) memory updator (see row w/o MemUpd).

We next examine whether our MRG outperforms MRC baselines in Table~\ref{tab:Results-for-bAbI}.
Generally, these baselines perform similar to the experiment on bAbI dataset \cite{Bordes2017LearningEG}.
Specifically, DMN+ is the strongest baseline, which achieves 43.97\% accuracy on average.
QRN, however, does not perform as well as it does on bAbI dataset, obtaining lower accuracy than DMN+.
Our model obtains highest mean accuracy and best accuracy over 10 runs among all baselins.

\textbf{Human evaluation.}
The results of human evaluation are listed in Table~\ref{tab:human_evaluation}. 
Our model significantly outperforms most of the baselines in terms of all the metrics. 
Particularly, our model increases informativeness approximately 4.1\% over PAC. 
This demonstrates that trying to answer reading comprehension question about dialog history if beneficial for improving and enriching the responses.

\subsection{Analysis of Multi-task learning}

Our  model  aims  to  generate response as well as answering MRC questions, which can be regarded as a multi-task.   
Hence,  in this subsection,  we  examine  whether  these  two  tasks can  complement  each  other.  
We list the performance on two single tasks by 'MRG w/o MT' in Table~\ref{tab:automated_evaluation} and Table~\ref{tab:Results-for-bAbI}, which solely generates response and answers MRC question, respectively.
It can be seen that by answering reading comprehension question, the performance of dialog generation increases by 12.1\% in terms of BLEU4 score, and by generating responses at the same time, MRC accuracy increases by 1.2\%.

\begin{table}
	\small
	\centering{\begin{tabular}{ccc}
			\toprule
			Model& Readability&Informativeness\\
			\midrule
			HRED&1.43& 1.46\\
			VAE& 1.60&1.58\\
			PAC& 1.56&1.72\\
			MRG& \textbf{1.63}&\textbf{1.75}\\
			\bottomrule
	\end{tabular}}
	\caption{Human evaluation on two aspects: Readability and informativeness. }
	\label{tab:human_evaluation}
\end{table}

\section{Conclusion}

In this paper we propose the multi-task framework to generate response and answer reading comprehension questions about multi-turn dialog.
Concretely, the two tasks share the same encoder to extract the common and task-invariant features with different decoders to learn specific features.
To better fusing information from the question and the dialog history in the encoding part, we propose to augment the Transformer architecture with a memory updater, which is designed to selectively store and update the history dialog information.
Experimental results show that our proposed model outperforms classic baselines.
In the future we would like to apply our model to other multi-task scenarios.

\section{Acknowledgments}
This work was supported by the National Key Research and Development Program of China (No. 2020YFB1406702), the National Science Foundation of China (NSFC No. 61876196 and No. 61672058), Beijing Outstanding Young Scientist Program No. BJJWZYJH012019100020098. Rui Yan is supported as a young fellow at Beijing Academy of Artificial Intelligence (BAAI).

\appendix
\bibliography{aaai21}
\end{document}